\RequirePackage[svgnames]{xcolor}

\documentclass[11pt,a4paper]{mystyle}

\usepackage[all]{hypcap}
\usepackage[comma,authoryear,compress]{natbib}
\usepackage{amsmath,amsfonts}
\usepackage{algorithmic}
\usepackage{algorithm}
\usepackage{array}
\usepackage{textcomp}
\usepackage{stfloats}
\usepackage{url}
\usepackage{verbatim}
\usepackage[font=small,labelfont=bf]{caption}
\usepackage{textcomp}
\usepackage{stfloats}
\usepackage{amssymb}
\usepackage{latexsym}
\usepackage[utf8]{inputenc}
\usepackage{microtype}
\usepackage{inconsolata}
\usepackage{ragged2e}
\usepackage{balance}
\usepackage{multirow}
\usepackage{multicol}
\usepackage{makecell}
\usepackage{tabularx,booktabs}
\usepackage{xspace}
\usepackage{fontawesome5}
\usepackage{float}
\usepackage{pifont}
\usepackage{microtype}
\usepackage{ulem}
\usepackage{bm}
\usepackage{autonum}
\usepackage{colortbl}
\usepackage{longtable}
\usepackage{hyperref}
\usepackage{wrapfig}
\usepackage{tcolorbox}
\usepackage{floatflt}
\usepackage{arydshln}
\usepackage{lineno}

\usepackage{dsfont}
\usepackage{cleveref}
\usepackage{bxcoloremoji}

\usepackage{tcolorbox}
\usepackage{multicol}
\usepackage{multirow}
\usepackage{tabularx}
\usepackage{adjustbox}
\usepackage{graphicx}
\usepackage{array}
\usepackage{color}
\usepackage{booktabs}
\usepackage{arydshln}
\usepackage{subfig}

\usepackage{bbding}

\usepackage{balance}
\usepackage{pifont}
\usepackage{array}
\usepackage{color}
\usepackage{booktabs}
\usepackage{arydshln}
\usepackage{rotating}
\definecolor{lightgrayv}{HTML}{F4F3F8} 
\definecolor{lightbluev}{HTML}{F5F9FD} 
\definecolor{grayv}{HTML}{707070}
\definecolor{bettergreen}{rgb}{0.13, 0.55, 0.13}
\definecolor{bluev}{HTML}{0070C0}
\definecolor{myblue}{HTML}{3889fe}
\definecolor{myorange}{HTML}{b2561a}
\definecolor{mygreen}{HTML}{6ea13f}
\definecolor{yellowv}{HTML}{d6a03d} 

\newcommand{\modelname}{\textsc{OmiGraph}\xspace}

\title{\textit{Reasoning About the Unsaid}:\\Misinformation Detection with Omission-Aware Graph Inference}

\runningtitle{Reasoning About the Unsaid: Misinformation Detection with Omission-Aware Graph Inference}

\author[1,2]{\href{https://scholar.google.com/citations?user=fMCBAt4AAAAJ}{\textcolor{black}{Zhengjia Wang}}}
\author[1]{\href{https://scholar.google.com/citations?user=hGZwK0cAAAAJ}{\textcolor{black}{Danding Wang}}}
\author[1]{\href{https://sheng-qiang.github.io/}{\textcolor{black}{Qiang Sheng}}}
\author[3]{\href{https://jiayingwu19.github.io/}{\textcolor{black}{Jiaying Wu}}}
\author[1,2]{\href{https://scholar.google.com/citations?user=fSBdNg0AAAAJ}{\textcolor{black}{Juan Cao}}}

\affil[1]{Media Synthesis and Forensics Lab, Institute of Computing Technology, Chinese Academy of Sciences}
\affil[2]{University of Chinese Academy of Sciences}
\affil[3]{National University of Singapore}

\correspondingauthor{Juan Cao}
\emails{
    {\{\href{mailto:wangzhengjia21b@ict.ac.cn}{wangzhengjia21b}, \href{mailto:wangdanding@ict.ac.cn}{wangdanding}, \href{mailto:shengqiang18z@ict.ac.cn}{shengqiang18z}\}@ict.ac.cn, \; \href{mailto:jiayingwu@u.nus.edu}{jiayingwu@u.nus.edu}, \; \href{mailto:caojuan@ict.ac.cn}{caojuan}@ict.ac.cn}
}

\begin{document}

\begin{abstract}
This paper investigates the detection of misinformation, which deceives readers by explicitly \textit{fabricating} misleading content or implicitly \textit{omitting} important information necessary for informed judgment. While the former has been extensively studied, omission-based deception remains largely overlooked, even though it can subtly guide readers toward false conclusions under the illusion of completeness.
To pioneer in this direction, this paper presents \modelname, the first omission-aware framework for misinformation detection. 
Specifically, \modelname constructs an omission-aware graph for the target news by utilizing a contextual environment that captures complementary perspectives of the same event, thereby surfacing potentially omitted contents.
Based on this graph, omission-oriented relation modeling is then proposed to identify the internal contextual dependencies, as well as the dynamic omission intents, formulating a comprehensive omission relation representation.
Finally, to extract omission patterns for detection, \modelname introduces omission-aware message-passing and aggregation that establishes holistic deception perception by integrating the omission contents and relations.
Experiments show that, by considering the omission perspective, our approach attains remarkable performance, achieving average improvements of $+5.4$\% F1 and $+5.3$\% ACC on two large-scale benchmarks.
\vspace{15pt}

\coloremojicode{1F4C5} \textbf{Date}: December 1, 2025

\coloremojicode{1F3E0} \textbf{Project}: \href{https://github.com/ICTMCG/OmiGraph}{https://github.com/ICTMCG/OmiGraph}

\coloremojicode{1F4AC} \textbf{Venue}: The 40th Annual AAAI Conference on Artificial Intelligence (AAAI 2026)
\end{abstract}

\maketitle

\begin{figure}[H]
\centering
\includegraphics[width=.6\linewidth]{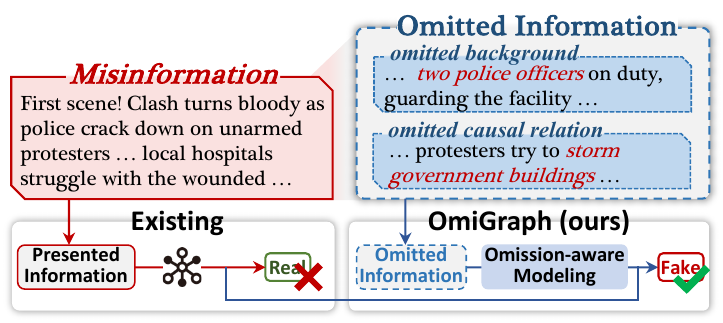}
\caption{\label{fig:motive}Concept comparison between existing methods and our \modelname. In contrast to solely focusing on the presented information, \modelname extracts more complete deception features via omission-aware modeling, capturing omitted key information such as the background and causal relation, and thus enhancing the existing detectors.}
\end{figure}

\section{Introduction}

The proliferation of misinformation poses severe societal consequences, from exacerbating conflicts to accelerating truth decay ~\citep{hu2025llm}. Therefore, developing effective automated detection methods has become a critical research objective \citep{wang2025bridging, wu2024fake}.

Misinformation is deceptive news strategically crafted to deceive readers~\citep{wang2025exploring, wu2025seeing, zhou2020survey}. Such deception manifests in two primary forms: (\textit{i}) the explicit \textit{fabrication} of specific narratives~\citep{greifeneder2020psychology}, and (\textit{ii}) the implicit \textit{omission} of critical information that is essential for comprehending the full context of an event and making informed judgments~\citep{chisholm1977intent, levine2022truth}. Prior research has predominantly focused on the former, detecting misinformation from ``what is said'', \textit{i.e.}, how information is presented, structured, and linguistically stylized to appear convincing or truthful. 
For example, state-of-the-art misinformation detection methods leverage explicit cues such as stylistic or emotional signals~\citep{zhang2021mining, xiao2024msynfd}, commonsense conflicts~\citep{wang2025robust}, or inconsistencies with external information~\citep{zheng2024evidence, yue2024retrieval} to identify deception.

However, omission-based deception remains largely underexplored despite its damaging role. Omissions are pervasive and often more insidious than explicit fabrications \citep{karlova2013social, van2014communicating}. Psychological studies \citep{turner1975information} also show that people are more likely to be deceived when information is selectively presented~\citep{pittarello2016legitimate, appling2015discriminative}.
As illustrated in \figurename~\ref{fig:motive}, an misinformation example concerning protests from \citep{sheng2022zoom} demonstrates how creators intentionally omit background information and causal relations to heighten perceived conflict between protesters and police, thereby facilitating deception.
Unfortunately, existing detection methods show limits in identifying such deceptive news, as they rely on explicit cues and fall short in capturing omission patterns that operate through what is deliberately left ``unsaid'' rather than what is explicitly stated.

Although detecting omission-based deception is essential for understanding the full spectrum of news deception, it presents three unique challenges:
(\textit{i}) \textit{Implicit signal recovery}: Omitted information is, by definition, absent from the target article and cannot be directly observed, making it hard to fill in the missing pieces.
(\textit{ii}) \textit{Dynamic omission relation}: The relation between what is stated and what is omitted varies widely based on the specific context, ranging from benign summarization to malicious manipulation such as obfuscating causality or distorting blame. Therefore, omission relation requires dynamic analysis, rather than predefining all such relation types.
(\textit{iii}) \textit{Omission pattern modeling}:  Extracting omission patterns of deception is challenging, as it involves not only effectively integrating omitted information with its corresponding relations, but also establishing holistic deception perception for misinformation detection.

To address these challenges, we propose \modelname, a pioneering omission-aware framework for misinformation detection.
First, drawing on the insight that news articles describing the same event often convey diverse yet complementary perspectives~\citep{Wang_EANN, sheng2022zoom}, we establish an omission-aware graph for the target news, which utilizes a contextual environment that leverages semantic similarity to help recover what might be left out.
Then, the contents from both sources (\textit{i.e.}, the target news and its contextual environment) are disassembled into fine-grained segment representations, serving as the graph nodes during initialization.

Second, based on the established graph, \modelname further proposes omission-oriented relation modeling, consisting of intra- and inter-source relation inferences, to formulate comprehensive representations of omission relations among graph nodes.
Specifically, intra-source relation focuses on identifying the contextual dependencies among segments within the same source, which reveal how internal segments interact to maintain narrative coherence or facilitate deception.
Besides, to capture dynamic omission intents among segments across different sources, inter-source relation modeling leverages large language models' (LLMs) superior capabilities in capturing implicit cues and contextual subtleties~\citep{yerukola2024pope, wang2025exploring}, instead of predefining all relation types.
Together with contextual dependencies, the resulting omission intents are transformed into edge attributes within the omission-aware graph, helping distinguish between legitimate editorial choices and potentially deceptive omissions. 

Finally, to extract omission patterns for detection, \modelname presents an omission-guided message passing and aggregation mechanism, which facilitates the utilization of omitted information for holistic deception perception. 
Concretely, it includes a local attention-based message passing that utilizes omission relations encoded in edge attributes to guide the propagation of omitted information among neighbor nodes, as well as a global aggregation that introduces a super root node to ensure a holistic understanding of omission patterns of deception across the entire narrative.

We evaluate \modelname on two large-scale English and Chinese datasets and achieve consistent performance gains, improving macro F1 scores by $2.91$--$17.03$\% in English and $2.97$--$13.44$\% in Chinese, respectively. 
Our results highlight the importance of \textit{reasoning about the unsaid} for misinformation detection, and we hope our findings will facilitate more omission-aware investigation in this field.

To sum up, our main contribution includes:
\begin{itemize}[leftmargin=*,itemsep=2pt,topsep=0pt,parsep=0pt]
    \item \textbf{Omission-aware detection framework}: 
    We propose the first omission-aware framework for misinformation detection that explicitly models omission-based deception.
    \item \textbf{Dynamic omission relation modeling}: 
    We develop a novel approach that enables dynamic inference of omission relations, together with omission-guided message passing and aggregation for holistic deception perception.
    \item \textbf{In-depth investigation}: 
    We validate the effectiveness across bilingual datasets, analyze omission patterns, and examine our approach's broader applicability.
\end{itemize}

\section{Related Works}
Existing misinformation detection methods can be categorized into three main categories based on the types of information requirements.
\textbf{(\textit{i}) Intrinsic information within the news itself.}
Early content-based approaches leverage linguistic and stylistic features, using handcrafted features such as syntax, sentiment, or lexical signals \citep{feng2012syntactic}. Deep learning methods further advance this direction by utilizing sentence-level or document-level embeddings and diverse features for improved performance \citep{wu2022probing, hu2023learn, wang2025bridging}.
For example, \citet{wang2025bridging} consider the intent conveyed in news pieces, incorporating intent signals for misinformation detection.
\textbf{(\textit{ii}) Collective wisdom from social context.} 
These methods utilize user profiles \citep{shu2019role, sitaula2020credibility}, user comments \citep{Shu_WeakSupervision, nan2024let}, and propagation structures \citep{sun2023fighting} for misinformation detection (\textit{e.g.}, a reply comment saying ``This is a known lie'' would be an important signal to make a prediction). 
For instance, \citet{zhao2025mppfnd} gather the propagation structure of a news claim across various social media platforms, analyze the characteristics of multi-platform news propagation, and explore how these differentiated propagation features aid in detecting misinformation.
\textbf{(\textit{iii}) External information for conflict verification.}
This group of methods relies on external evidence \citep{zheng2024evidence, yue2024retrieval, wu2025beyond} or commonsense \citep{wang2025robust, hu2024bad} to identify the inconsistencies of what is said and ground facts.
For example, \citet{yue2024retrieval} utilize LLMs to form contrastive arguments conditioned on the evidence from varying perspectives, then incorporates nuanced information for LLM-based verification. \citet{wang2025robust} design a commonsense conflict detection framework, utilize an external commonsense tool to help detect the inconsistencies, providing signals for misinformation detection.

However, these approaches focus on the commission patterns of presented news content, \textit{i.e.}, ``what is said'', while overlooking the more subtle \textit{omission patterns} of deception through strategic exclusion of crucial information \citep{carson2010lying, van2022effects}.
In contrast, our proposed \modelname introduces the first omission-aware framework that explicitly models the omitted information and intent, effectively enhancing existing misinformation detectors.

\begin{figure*}[t]
\centering
    \includegraphics[width=1.\linewidth]{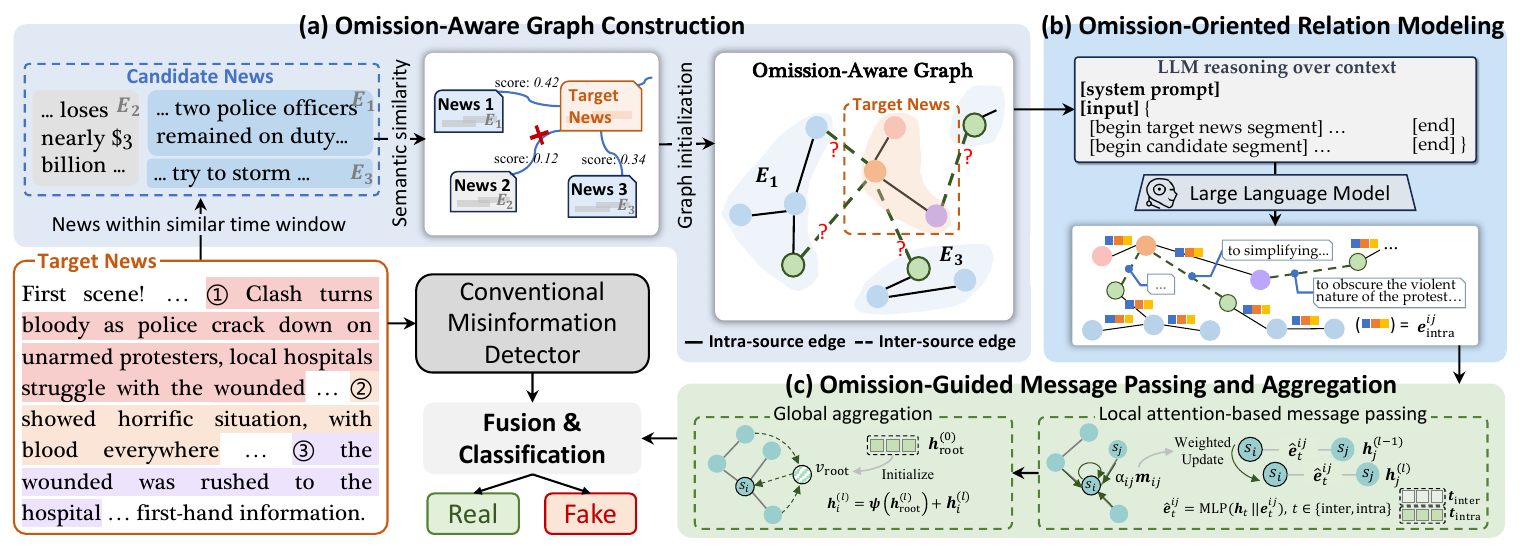}
    \caption{\label{fig:method} Overview of \modelname. Given a news piece, \modelname constructs an omission-aware graph based on the contextual environment (a). Then, omission-oriented relation modeling reasons over the graph nodes, identifying intra-source contextual dependencies and inter-source omission intents (b). Finally, an omission-guided message passing mechanism extracts omission-oriented deception features (c) to enhance conventional misinformation detectors.}
\end{figure*}

\section{Proposed Method: \modelname}
To capture omission patterns of deception, we propose the first omission-aware misinformation detection framework, \modelname. 
As illustrated in Figure~\ref{fig:method}, \modelname consists of three key components: omission-aware graph construction, omission-oriented relation modeling, and omission-guided message passing and aggregation. Through these proposed modules, \modelname is able to extract omission patterns of deception, thus significantly enhancing conventional misinformation detection methods.

\subsection{Omission-aware Graph Construction}
To effectively leverage the omitted information for misinformation detection, \modelname establishes an omission-aware graph architecture, utilizing a contextual environment that serves as the resource for recovering information omitted from the target news.

\noindent \textbf{Contextual environment construction.}
Motivated by the observation that contemporary news articles covering related topics naturally offer diverse perspectives for each other \citep{Wang_EANN, sheng2022zoom}, \modelname constructs a corresponding contextual environment to assist in recovering the omitted information.
Specifically, we leverage the semantic similarity to select potential contextual news with similar time and relevant topics.
Given a target news item $n_\text{tgt}$ published at time $T_\text{tgt}$, a candidate news pool $\mathcal{P}$ containing articles published $T$ days (\textit{e.g.}, $T=3$) before $n_\text{tgt}$ is constructed:
\begin{equation}
    \mathcal{P} = \{n_{p} \;|\; T_{p} \in [T_\text{tgt} - T, \; T_\text{tgt})\}.
\end{equation}
Then, to identify news articles sharing a relevant topic, a pre-trained language model (\textit{e.g.}, BERT) is employed for semantic similarity computation.
For each candidate item $n_u$ in $\mathcal{P}$, we leverage BERT to obtain its coarse semantic representations, denoted as $\mathbf{h}_{u}$, and rank all candidates in $\mathcal{P}$ by their cosine similarity to the ones of the target news $\mathbf{h}_\text{tgt}$.
Finally, the top-$K$ similar items are selected as the corresponding contextual environment $\mathcal{C}_\text{ctx}$, as follows:
\begin{equation}
    \mathcal{C}_\text{ctx} = \{n_{u} \;|\; n_{u} \in \text{TopK}\left(\text{cos}(\mathbf{h}_\text{tgt}, \mathbf{h}_{u}), \; n_u \in \mathcal{P}\right)\},
\end{equation}
where cosine similarity $\text{cos}(\cdot)$ is calculated as $\text{cos}(\mathbf{a}, \mathbf{b}) = \frac{\mathbf{a} \cdot \mathbf{b}}{||\mathbf{a}||_2 \cdot ||\mathbf{b}||_2}$, and $\mathbf{a}, \mathbf{b}$ are feature vectors.
In this way, \modelname can utilize the constructed contextual environment as a valuable resource for subsequently recovering the content part of omitted information.

\noindent \textbf{Omission-aware graph initialization.}
Rather than directly adopting the entire news representations (\textit{i.e.}, $\mathbf{h}_{u}$ and $\mathbf{h}_\text{tgt}$) as the graph nodes, to enable the fine-grained omission analysis, \modelname further disassembles the entire target news and its contextual environment into sentence-level atomic segments, thus facilitating the precise identification of omission boundaries.
Concretely, the omission-aware graph can be formalized in $\mathcal{G} = \{\mathcal{V}, \mathcal{E}\}$ with content segment representations as nodes $\mathcal{V}$, and different relations among them as edges $\mathcal{E}$, as follows:
\begin{equation}
\begin{aligned}
    \mathcal{V} &= \left\{\mathbf{h}_{u}^{i} \;|\; n_u \in \{n_\text{tgt}\} \cup \mathcal{C}_\text{ctx}\right\},\\
    \mathcal{E} &= \left\{\mathcal{E}_{\text{intra}}, \; \mathcal{E}_{\text{inter}}\right\},
\end{aligned}
\end{equation}
where $\mathbf{h}_{u}^{i}$ represents the encoded $i$-th segments of news $n_{u}$. The graph edges consist of intra-source edges ($\mathcal{E}_{\text{intra}}$) connecting nodes within the same source, as well as inter-source edges ($\mathcal{E}_{\text{inter}}$) across the target news $n_\text{tgt}$ and its contextual environment $\mathcal{C}_\text{ctx}$, which contains a total of $k$ segments.
This design enables \modelname to explicitly model both fine-grained content representations and specific relations.

\subsection{Omission-oriented Relation Modeling}
Based on the initialized nodes $\mathcal{V}$, \modelname further presents omission-oriented relation modeling to effectively extract the comprehensive relations among them, consisting of intra-source relations for internal contextual dependencies and inter-source relations for dynamic omission intents.

\noindent \textbf{Intra-source relation.}
To identify the contextual dependencies among segments within the same news article, we introduce intra-source relations for modeling semantic consistency patterns. This captures how segments interact to maintain narrative coherence or facilitate deception.

Specifically, for each news item in the constructed graph $\mathcal{G}$, we introduce learnable edge embeddings that act as semantic bridges between its internal segments, modeling both individual segment characteristics and their interaction patterns through intra-source edges:
\begin{equation}
    \begin{aligned}
        \mathbf{e}_\text{intra}^{ij} = \mathrm{MLP}\left(\mathbf{h}_{i} \;\|\; \mathbf{h}_{j} \;\|\; \mathrm {diff}(\mathbf{h}_{i} - \mathbf{h}_{j})\right), 
    \end{aligned}
\end{equation}
where $\mathbf{h}_{i}$ and $\mathbf{h}_{j}$ represent the encoded embeddings of the two segments within the same news item, respectively. Here, $\mathrm{MLP}(\cdot)$ is a multi-layer perceptron, $\|$ denotes the vector concatenation, and $\mathrm{diff}(\cdot)$ computes the element-wise absolute difference.

These learnable edge embeddings constitute the attributes of intra-source edges that establish adaptive semantic bridges within individual news articles, enabling the discovery of subtle contextual patterns:
\begin{equation}
    \mathcal{E}_{\text{intra}} = \{ \mathbf{e}_\text{intra}^{ij} \;|\; \mathbf{h}_{i}, \mathbf{h}_{j} \in \mathcal{V}\}.
\end{equation}

\noindent \textbf{Inter-source relation.}
To capture dynamic omission intents among segments across different sources, instead of attempting to predefine all possible relation types, a more flexible strategy is proposed for inter-source relation inference. 
Beyond binary relation detection (\textit{i.e.}, whether omission relation exists), omission intent reasoning is performed between target and contextual news segments to understand motivations and potential deceptive impact of such omissions.

Concretely, considering that identifying omission intents inherently requires understanding nuanced context and implicit intent---capabilities where LLMs excel~\citep{yerukola2024pope, wang2025exploring}---we employ LLM assistance to reason about underlying omission intents, such as benign summarization or malicious manipulation. 
Given the omission-aware graph $\mathcal{G}$, we elicit reasoning from an LLM $\mathcal{M}$ to analyze pairs of target and contextual news segments, inferring whether an omission relation exists and if so, \textit{why} information from contextual news segments is intentionally omitted:
\begin{equation}
\begin{aligned}
    \mathbf{e}_\text{inter}^{ij} &= \mathrm{PLM}\Big(\mathcal{M}(s_\text{tgt}^{i}, \; s_\text{ctx}^{j})\Big),\\
    \mathcal{E}_{\text{inter}} &= \{ \mathbf{e}_\text{inter}^{ij} \;|\; s_\text{tgt}^i \in n_\text{tgt},\; s_\text{ctx}^{j} \in n_\text{ctx} \in \mathcal{C}_\text{ctx}\},
\end{aligned}  
\end{equation}
where $s_\text{tgt}^{i}$ and $s_\text{ctx}^{j}$ represent the $i$-th segment text of target news and $j$-th segment of contextual news within the environment, respectively.
In practice, $\mathcal{M}(\cdot)$ returns omission intents between pairs of inter-source news segments through free-text descriptions (\textit{e.g.}, ``to downplay the political motivations behind actions''), which are then encoded into edge attributes using a pre-trained language model $\mathrm{PLM}(\cdot)$.

Through the above process, we can transform dynamic omission intents into interpretable text-attributed edge representations accessible to subsequent graph learning, enhancing both relational expressiveness and interpretability.

\subsection{Omission-guided Message Passing and Aggregation}
To model omission patterns of deception, \modelname introduces omission-guided message passing and aggregation, including local attention-based message passing that leverages omission relations to guide omitted information propagation and a global aggregation that enables a holistic deception representation for misinformation detection.

\noindent\textbf{Local attention-based message passing.}
Based on the constructed $\mathcal{G}$, local attention-based message passing leverages the valuable omission-oriented relation encoded in both intra- and inter-source edges to guide the transmission of omitted information within the omission-aware graph.

To distinguish relation types and their semantic roles in the omission-aware graph, we enhance edge representations by incorporating learnable type-specific embeddings. Concretely, for the edge between different nodes, its enhanced attributes are calculated as:
\begin{equation}
    \mathbf{\hat{e}}_{t}^{ij} = \mathrm{MLP}\left(\mathbf{h}_{t} \;\|\; \mathbf{e}_{t}^{ij}\right), \; t \in \{\text{intra}, \text{inter}\},
\end{equation}
where $\mathbf{h}_{t}$ is a learnable type-specific embedding, representing the type (\textit{i.e.}, intra-source or inter-source) of the current edge, and $\mathbf{e}_{t}^{ij}$ is the original edge attribute derived from the corresponding relation modeling process.

Finally, the local message passing operation employs an attention mechanism to selectively aggregate neighborhood information. 
Concretely, given any one node $\mathbf{h}_{i} \in \mathcal{V}$, its representation at layer $l$ is updated from the previous layer:
\begin{equation}
\begin{aligned}
    \mathbf{h}^{(l)}_i = &\text{MLP}\Big(\mathbf{h}^{(l-1)}_i + \sum_{\mathbf{h}_j \in \mathcal{N}(\mathbf{h}_i)} \alpha_{ij} \cdot \mathbf{m}_{ij}\Big), \\
    &\mathbf{m}_{ij} = \mathrm{MLP}\big(\mathbf{h}^{(l-1)}_j \;\|\; \mathbf{\hat{e}}_{t}^{ij}\big),
\end{aligned}
\end{equation}
where $\mathbf{h}_j \in \mathcal{N}(\mathbf{h}_i)$ represents the neighbor nodes of current $\mathbf{h}_i$, and $\mathbf{m}_{ij}$ incorporates both neighbor node information and edge information.
And the attention weight $\alpha_{ij}$ is computed as follows:
\begin{equation}
    \alpha_{ij} = \text{softmax}\left((\mathbf{h}_i^{(l-1)} + \mathbf{\hat{e}}_{t}^{ij}) \cdot (\mathbf{h}_j^{(l-1)} + \mathbf{\hat{e}}_{t}^{ij})\right).
\end{equation}
This attention-based message passing mechanism enables \modelname to leverage edge-encoded omission relations for selective information propagation, facilitating the recovery and integration of omitted information across the graph.

\noindent\textbf{Global aggregation.}
Purely adopting local information message passing is limited by its inability to model global coherence, which is crucial for understanding holistic deceptive patterns. 
Although stacking multi-layer local message passing can achieve a larger receptive field, it may result in over-smoothing \citep{oonograph} and over-squashing \citep{alonbottleneck} that dilute critical omission signals.
Therefore, to obtain a holistic understanding of omission-based deception across the entire narrative, a global aggregation strategy is incorporated by introducing a super root node that serves as a central aggregator for global omission-aware information. 

Concretely, the super root node is implemented as a learnable embedding $\mathbf{h}_{\text{root}}$ which is randomly initialized and optimized during training.
At layer $l$, the super root node aggregates weighted information from all graph nodes:
\begin{equation}
    \mathbf{h}^{(l)}_{\text{root}} = \mathbf{h}^{(l-1)}_{\text{root}} + \sum_{i} \mathrm{softmax}(\mathbf{W}\mathbf{h}_{i}^{(l-1)} + b) \cdot \mathbf{h}_{i}^{(l-1)}, 
\end{equation}
where $\mathbf{h}_{i}^{(l-1)}$ represents the embedding of any one node $\mathbf{h}_{i}\in \mathcal{V}$ at the ($l-1$)-th layer, and $\mathbf{W}$ and $b$ are learnable parameters that determine each node's contribution to the global narrative understanding.
The updated global information is then integrated back into individual node representations through residual fusion:
\begin{equation}
    \mathbf{h}_{i}^{(l)} \leftarrow \psi(\mathbf{h}^{(l)}_{\text{root}}) + \mathbf{h}_{i}^{(l)},
\end{equation}
where $\psi(\cdot)$ is a non-linear transformation. 
This global-local integration ensures that segment-level omission patterns are contextualized within the overall narrative structure, enabling the model to distinguish between localized information gaps and systematic omission-based deception.

\subsection{Model Prediction and Optimization}
To perform misinformation detection, we aggregate features from $\mathcal{G}$ by applying mean pooling over the node embeddings of the target news.
The pooled representation $\mathbf{h}_\text{omi}$ is then fused with conventional commission-based misinformation detection signals to generate the final prediction $\hat{y}\in[0,1]$:
\begin{equation}
\hat{y} = \mathrm{fuse}(\mathbf{h}_\text{omi} \;\|\; \mathbf{h}_\text{com}),
\end{equation}
where $\mathbf{h}_\text{com}$ represents features from commission-based detectors or general text encoders (e.g., BERT), and $\mathrm{fuse}(\cdot)$ represents the fusion mechanism.

Finally, the binary cross-entropy loss is utilized to optimize the model parameters:
\begin{equation}
\mathcal{L}_\text{cls} = - y \log(\hat{y}) - (1 - y) \log(1 - \hat{y}),
\end{equation}
where $y \in \{0, 1\}$ is ground-truth label.

\section{Experiments} \label{sec:exp}
In this section, we present empirical results to demonstrate the effectiveness of \modelname. 

\subsection{Experimental Setup}

\subsubsection{Datasets}

We evaluate the proposed \modelname on two public datasets from \citep{sheng2022zoom}, which enables cross-lingual evaluation. 
The \textit{English} dataset comprises verified posts from Twitter (now X) and fact-checking websites. 
The \textit{Chinese} dataset consists of posts collected from Weibo, China's major social media platform. 
Both datasets are coupled with contemporary media coverage to establish comprehensive contextual news corpora, including 1,003,646 and 583,208 contemporaneous news articles for the English and Chinese datasets, respectively. 
Detailed information can be found in the Appendix.

\subsubsection{Baselines} 
Technically, our \modelname could coordinate with any misinformation detectors. Specifically, we include two groups of existing methods for comparison.
The first group includes content-only detection methods:
\begin{itemize}[leftmargin=*,itemsep=2pt,topsep=0pt,parsep=0pt]
    \item \textit{BERT} \citep{devlin_BERT}, a pre-trained language model widely used as the text encoder for misinformation detection~\cite{zhu2022generalizing, xiao2024msynfd}, with the last layer finetuned conventionally. 
    \item \textit{DualEmo} \citep{zhang2021mining}, considering the emotions conveyed in news pieces for misinformation detection.
    \item \textit{MSynFD} \citep{xiao2024msynfd}, a structure-aware model that builds a multi-hop syntactic dependency graph to model syntax information and sequentially aware semantic information for misinformation detection.
    \item  \textit{LLM}, to validate the performance of LLM in the misinformation detection task, we prompt an LLM to make veracity judgments based on the provided news content.
    \item  \textit{PCoT} \citep{modzelewski2025pcot}, models misinformation from persuasion, integrating persuasion knowledge into the reasoning of LLMs for misinformation detection.
\end{itemize}
The second group includes conflict-aware methods based on external information:
\begin{itemize}[leftmargin=*,itemsep=2pt,topsep=0pt,parsep=0pt]
    \item \textit{NEP}~\citep{sheng2022zoom}, leverages concurrent mainstream media news to model uniqueness and popularity features of target news for misinformation detection.
    \item \textit{MD-PCC}~\citep{wang2025robust}, utilizes an external commonsense tool to detect commonsense conflicts within news content, providing inconsistency signals.
    \item \textit{RAV}~\citep{zheng2024evidence}, an end-to-end enhanced evidence selection-based news verification method.
    \item \textit{RAFTS}~\citep{yue2024retrieval}, employs LLMs to construct contrastive arguments based on evidence, enabling nuanced reasoning for verification.
\end{itemize}
To ensure fairness, all LLM-based methods employ the same model. 
Evidence-based methods are implemented following \citep{sheng2022zoom}, considering the verification results as the misinformation detection results.
Implementation details are provided in the Appendix due to space constraints.

\begin{sidewaystable} \scriptsize
    \centering
    \renewcommand{\arraystretch}{1.3}
    \setlength\tabcolsep{4.5pt}
    \caption{Performance comparison of base models with and without \modelname. The better results in each group using the same base model are \textbf{bolded}. The {$\pm$} denotes the standard deviation, * indicates 0.005 significance level from a paired t-test comparing \modelname with its base model.}
    \begin{tabular}{cl llll llll}
    \toprule
    \multicolumn{2}{c}{\multirow{2}{*}[-0.3em]{\textbf{Method}}} 
    & \multicolumn{4}{c}{\textbf{Dataset: \textit{English}}} & \multicolumn{4}{c}{\textbf{Dataset: \textit{Chinese}}} 
    \\
    \cmidrule(lr){3-6} \cmidrule(lr){7-10}
    & & \makecell[c]{macF1} & \makecell[c]{Acc} & \makecell[c]{F1$_{\text{real}}$} & \makecell[c]{F1$_{\text{fake}}$}
    & \makecell[c]{macF1} & \makecell[c]{Acc} & \makecell[c]{F1$_{\text{real}}$} & \makecell[c]{F1$_{\text{fake}}$}
    \\
    \midrule
    \multirow{10}{*}{\rotatebox{90}{\makecell{Content-only}}} 
        & BERT 
            & 0.7111{\scriptsize \color{grayv} $\pm$.0032} 
            & 0.7135{\scriptsize \color{grayv} $\pm$.0021} 
            & 0.7367{\scriptsize \color{grayv} $\pm$.0035}
            & 0.7025{\scriptsize \color{grayv} $\pm$.0097} 

            & 0.7851{\scriptsize \color{grayv} $\pm$.0014} 
            & 0.7921{\scriptsize \color{grayv} $\pm$.0016} 
            & 0.8240{\scriptsize \color{grayv} $\pm$.0018} 
            & 0.7461{\scriptsize \color{grayv} $\pm$.0012}
        \\
        & \; + Ours
            & \textbf{0.7530}{\scriptsize \color{grayv} $\pm$.0017}* 
            & \textbf{0.7532}{\scriptsize \color{grayv} $\pm$.0016}* 
            & \textbf{0.7541}{\scriptsize \color{grayv} $\pm$.0077}* 
            & \textbf{0.7519}{\scriptsize \color{grayv} $\pm$.0109}* 
            
            & \textbf{0.8407}{\scriptsize \color{grayv} $\pm$.0036}* 
            & \textbf{0.8426}{\scriptsize \color{grayv} $\pm$.0040}* 
            & \textbf{0.8561}{\scriptsize \color{grayv} $\pm$.0098}* 
            & \textbf{0.8230}{\scriptsize \color{grayv} $\pm$.0018}* 
        \\
        
        & DualEmo
            & 0.7194{\scriptsize \color{grayv} $\pm$.0024} 
            & 0.7200{\scriptsize \color{grayv} $\pm$.0021} 
            & 0.7322{\scriptsize \color{grayv} $\pm$.0013} 
            & 0.7065{\scriptsize \color{grayv} $\pm$.0048} 

            & 0.7958{\scriptsize \color{grayv} $\pm$.0033} 
            & 0.8003{\scriptsize \color{grayv} $\pm$.0029} 
            & 0.8262{\scriptsize \color{grayv} $\pm$.0024} 
            & 0.7655{\scriptsize \color{grayv} $\pm$.0056} 
        \\     
        & \; + Ours
            & \textbf{0.7557}{\scriptsize \color{grayv} $\pm$.0003}* 
            & \textbf{0.7563}{\scriptsize \color{grayv} $\pm$.0006}* 
            & \textbf{0.7650}{\scriptsize \color{grayv} $\pm$.0000}* 
            & \textbf{0.7456}{\scriptsize \color{grayv} $\pm$.0000}*

            & \textbf{0.8417}{\scriptsize \color{grayv} $\pm$.0019}* 
            & \textbf{0.8433}{\scriptsize \color{grayv} $\pm$.0029}* 
            & \textbf{0.8584}{\scriptsize \color{grayv} $\pm$.0054}* 
            & \textbf{0.8236}{\scriptsize \color{grayv} $\pm$.0007}* 
        \\
        
        & MSynFD
             & 0.7317{\scriptsize \color{grayv} $\pm$.0018} 
            & 0.7319{\scriptsize \color{grayv} $\pm$.0017} 
            & 0.7324{\scriptsize \color{grayv} $\pm$.0103} 
            & 0.7309{\scriptsize \color{grayv} $\pm$.0083} 
            
            & 0.8054{\scriptsize \color{grayv} $\pm$.0052} 
            & 0.8089{\scriptsize \color{grayv} $\pm$.0048} 
            & 0.8315{\scriptsize \color{grayv} $\pm$.0034} 
            & 0.7793{\scriptsize \color{grayv} $\pm$.0074}
        \\
        & \; + Ours
            & \textbf{0.7608}{\scriptsize \color{grayv} $\pm$.0007}* 
            & \textbf{0.7647}{\scriptsize \color{grayv} $\pm$.0011}* 
            & \textbf{0.7586}{\scriptsize \color{grayv} $\pm$.0111}* 
            & \textbf{0.7528}{\scriptsize \color{grayv} $\pm$.0093}*

            & \textbf{0.8496}{\scriptsize \color{grayv} $\pm$.0061}* 
            & \textbf{0.8507}{\scriptsize \color{grayv} $\pm$.0055}* 
            & \textbf{0.8668}{\scriptsize \color{grayv} $\pm$.0046}* 
            & \textbf{0.8361}{\scriptsize \color{grayv} $\pm$.0103}*
        \\

        & LLM
            & 0.5556{\scriptsize \color{grayv} $\pm$.0002} 
            & 0.5779{\scriptsize \color{grayv} $\pm$.0000} 
            & 0.4561{\scriptsize \color{grayv} $\pm$.0002} 
            & 0.6552{\scriptsize \color{grayv} $\pm$.0013}

            & 0.6992{\scriptsize \color{grayv} $\pm$.0001} 
            & 0.7110{\scriptsize \color{grayv} $\pm$.0001} 
            & 0.6397{\scriptsize \color{grayv} $\pm$.0001} 
            & 0.7588{\scriptsize \color{grayv} $\pm$.0001}
        \\        
        & \; + Ours
            & \textbf{0.7259}{\scriptsize \color{grayv} $\pm$.0001}* 
            & \textbf{0.7305}{\scriptsize \color{grayv} $\pm$.0001}* 
            & \textbf{0.7610}{\scriptsize \color{grayv} $\pm$.0001}* 
            & \textbf{0.6908}{\scriptsize \color{grayv} $\pm$.0002}* 
            
            & \textbf{0.8336}{\scriptsize \color{grayv} $\pm$.0001}* 
            & \textbf{0.8367}{\scriptsize \color{grayv} $\pm$.0002}* 
            & \textbf{0.8563}{\scriptsize \color{grayv} $\pm$.0002}* 
            & \textbf{0.8109}{\scriptsize \color{grayv} $\pm$.0001}*
        \\

        & PCoT
            & 0.6508{\scriptsize \color{grayv} $\pm$.0011} 
            & 0.6509{\scriptsize \color{grayv} $\pm$.0001} 
            & 0.6434{\scriptsize \color{grayv} $\pm$.0000} 
            & 0.6481{\scriptsize \color{grayv} $\pm$.0003}

            & 0.8020{\scriptsize \color{grayv} $\pm$.0001} 
            & 0.8041{\scriptsize \color{grayv} $\pm$.0002} 
            & 0.7812{\scriptsize \color{grayv} $\pm$.0001} 
            & 0.8227{\scriptsize \color{grayv} $\pm$.0001}
        \\
        & \; + Ours
            & \textbf{0.7062}{\scriptsize \color{grayv} $\pm$.0003}* 
            & \textbf{0.7196}{\scriptsize \color{grayv} $\pm$.0001}* 
            & \textbf{0.7649}{\scriptsize \color{grayv} $\pm$.0000}* 
            & \textbf{0.6563}{\scriptsize \color{grayv} $\pm$.0001}* 
            
            & \textbf{0.8383}{\scriptsize \color{grayv} $\pm$.0001}* 
            & \textbf{0.8414}{\scriptsize \color{grayv} $\pm$.0002}* 
            & \textbf{0.8607}{\scriptsize \color{grayv} $\pm$.0001}* 
            & \textbf{0.8158}{\scriptsize \color{grayv} $\pm$.0002}*
        \\
        
        \midrule
        \multirow{8}{*}{\rotatebox{90}{\makecell{External\\information-aware}}}
        & NEP
            & 0.7274{\scriptsize \color{grayv} $\pm$.0004} 
            & 0.7278{\scriptsize \color{grayv} $\pm$.0005} 
            & 0.7383{\scriptsize \color{grayv} $\pm$.0011} 
            & 0.7165{\scriptsize \color{grayv} $\pm$.0005}

            & 0.8288{\scriptsize \color{grayv} $\pm$.0010} 
            & 0.8311{\scriptsize \color{grayv} $\pm$.0010} 
            & 0.8486{\scriptsize \color{grayv} $\pm$.0012} 
            & 0.8090{\scriptsize \color{grayv} $\pm$.0017} 
        \\ 
        & \; + Ours
            & \textbf{0.7596}{\scriptsize \color{grayv} $\pm$.0014}* 
            & \textbf{0.7647}{\scriptsize \color{grayv} $\pm$.0011}* 
            & \textbf{0.7586}{\scriptsize \color{grayv} $\pm$.0111}* 
            & \textbf{0.7528}{\scriptsize \color{grayv} $\pm$.0093}*             
            
            & \textbf{0.8585}{\scriptsize \color{grayv} $\pm$.0072}* 
            & \textbf{0.8596}{\scriptsize \color{grayv} $\pm$.0065}* 
            & \textbf{0.8711}{\scriptsize \color{grayv} $\pm$.0077}*
            & \textbf{0.8460}{\scriptsize \color{grayv} $\pm$.0061}*
        \\

        & MD-PCC
            & 0.7227{\scriptsize \color{grayv} $\pm$.0028} 
            & 0.7243{\scriptsize \color{grayv} $\pm$.0031} 
            & 0.7434{\scriptsize \color{grayv} $\pm$.0048} 
            & 0.7021{\scriptsize \color{grayv} $\pm$.0017}
            
            & 0.8168{\scriptsize \color{grayv} $\pm$.0022} 
            & 0.8205{\scriptsize \color{grayv} $\pm$.0026} 
            & 0.8427{\scriptsize \color{grayv} $\pm$.0033} 
            & 0.7909{\scriptsize \color{grayv} $\pm$.0019}
        \\
        & \; + Ours
            & \textbf{0.7572}{\scriptsize \color{grayv} $\pm$.0027}* 
            & \textbf{0.7593}{\scriptsize \color{grayv} $\pm$.0055}* 
            & \textbf{0.7650}{\scriptsize \color{grayv} $\pm$.0030}* 
            & \textbf{0.7456}{\scriptsize \color{grayv} $\pm$.0076}*
            
            & \textbf{0.8564}{\scriptsize \color{grayv} $\pm$.0081}* 
            & \textbf{0.8577}{\scriptsize \color{grayv} $\pm$.0075}* 
            & \textbf{0.8700}{\scriptsize \color{grayv} $\pm$.0086}* 
            & \textbf{0.8427}{\scriptsize \color{grayv} $\pm$.0098}* 
        \\

        & RAV
            & 0.7189{\scriptsize \color{grayv} $\pm$.0020} 
            & 0.7197{\scriptsize \color{grayv} $\pm$.0019} 
            & 0.7336{\scriptsize \color{grayv} $\pm$.0037} 
            & 0.7041{\scriptsize \color{grayv} $\pm$.0050}

            & 0.7930{\scriptsize \color{grayv} $\pm$.0018} 
            & 0.7980{\scriptsize \color{grayv} $\pm$.0011} 
            & 0.8252{\scriptsize \color{grayv} $\pm$.0016} 
            & 0.7608{\scriptsize \color{grayv} $\pm$.0047} 
        \\ 
        & \; + Ours
            & \textbf{0.7433}{\scriptsize \color{grayv} $\pm$.0056}* 
            & \textbf{0.7435}{\scriptsize \color{grayv} $\pm$.0056}* 
            & \textbf{0.7486}{\scriptsize \color{grayv} $\pm$.0076}* 
            & \textbf{0.7381}{\scriptsize \color{grayv} $\pm$.0067}* 
            
            & \textbf{0.8354}{\scriptsize \color{grayv} $\pm$.0049}* 
            & \textbf{0.8367}{\scriptsize \color{grayv} $\pm$.0041}* 
            & \textbf{0.8555}{\scriptsize \color{grayv} $\pm$.0036}*
            & \textbf{0.8184}{\scriptsize \color{grayv} $\pm$.0063}*
        \\   

        & RAFTS
            & 0.6016{\scriptsize \color{grayv} $\pm$.0005} 
            & 0.6049{\scriptsize \color{grayv} $\pm$.0006} 
            & 0.6019{\scriptsize \color{grayv} $\pm$.0010} 
            & 0.6208{\scriptsize \color{grayv} $\pm$.0009}

            & 0.7427{\scriptsize \color{grayv} $\pm$.0003} 
            & 0.7580{\scriptsize \color{grayv} $\pm$.0002} 
            & 0.8055{\scriptsize \color{grayv} $\pm$.0007} 
            & 0.6800{\scriptsize \color{grayv} $\pm$.0004} 
        \\
        & \; + Ours
            & \textbf{0.6771}{\scriptsize \color{grayv} $\pm$.0013}* 
            & \textbf{0.6907}{\scriptsize \color{grayv} $\pm$.0008}* 
            & \textbf{0.7381}{\scriptsize \color{grayv} $\pm$.0004}* 
            & \textbf{0.6401}{\scriptsize \color{grayv} $\pm$.0022}* 

            & \textbf{0.7870}{\scriptsize \color{grayv} $\pm$.0010}* 
            & \textbf{0.7928}{\scriptsize \color{grayv} $\pm$.0007}* 
            & \textbf{0.8222}{\scriptsize \color{grayv} $\pm$.0004}* 
            & \textbf{0.7517}{\scriptsize \color{grayv} $\pm$.0005}* 
        \\
    \bottomrule
    \end{tabular}
    \label{tab:main_result}
\end{sidewaystable}

\subsection{Overall Performance Comparison}

The average results over three runs are shown in \tablename~\ref{tab:main_result}. Experimental results show that: 
(\textit{i}) Consistent improvements across baselines. \modelname enhances various methods by $2.91$--$17.03$\% on English and $2.97$--$13.44$\% on Chinese in macro F1 scores, demonstrating that our omission-aware framework captures previously overlooked deceptive patterns and provides benefits to existing mechanisms.
(\textit{ii}) Effectiveness against LLM-based methods. \modelname delivers substantial gains even over advanced language models, showing our framework provides value beyond their inherent understanding of textual patterns.
(\textit{iii}) Complementary benefits to external information-aware methods. \modelname enhances the performance of methods that already utilize external information sources, including verification-based approaches, by identifying information completeness gaps rather than factual contradictions.

\subsection{Effectiveness of \modelname's Design}

\begin{figure}
    \centering 
    \includegraphics[width=.7\linewidth]{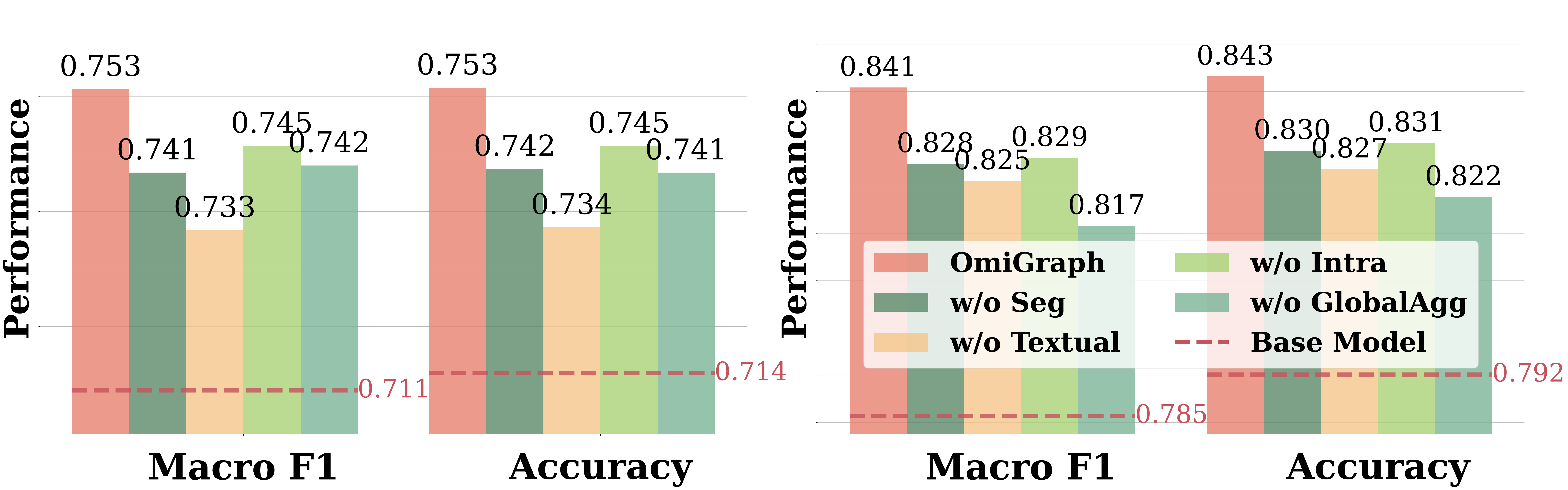}
    \caption{\label{fig:ablation} Performance comparison of \modelname and its variants using BERT as the base detector on the \textit{English} (left) and \textit{Chinese} (right) datasets.}
\end{figure}

To investigate the contribution of each key component in \modelname, we conduct the following studies:
\begin{itemize}[leftmargin=*,itemsep=2pt,topsep=0pt,parsep=0pt]
    \item \modelname (\textit{w/o} Seg), which skips the decomposition of news into fine-grained information segments and instead treats each news item as a single node.
    \item \modelname (\textit{w/o} Textual), which replaces omission intent textual edges with uniform structural connections between environmental and target news segments.
    \item \modelname (\textit{w/o} Intra), which removes intra-source connections, retaining only inter-source connections that represent omitted information.
    \item \modelname (\textit{w/o} GlobalAgg), which removes the global-level aggregation module, and performs only local message passing in the graph learning phase.
\end{itemize}
For clean comparison, we use BERT as the base misinformation detector across all variants. As shown in \figurename~\ref{fig:ablation}, the results reveal that:  
(\textit{i}) Fine-grained segment-level representation (\textit{w/o} Seg) enables more precise alignment for omitted information modeling; removing it results in coarse representations that hinder reasoning.
(\textit{ii}) Omission-oriented reasoning through LLM-generated textual edges (\textit{w/o} Textual) plays a crucial role in capturing implicit omission relations beyond structural or semantic proximity; its removal results in limited detection of subtle omission patterns.
(\textit{iii}) Intra-source relation (\textit{w/o} Intra) enhances the model’s ability to infer omissions by capturing contextual dependencies and providing enriched contextual cues. Their removal leads to weakened modeling.
(\textit{iv}) Global aggregation (\textit{w/o} GlobalAgg) facilitates the modeling of systematic omission patterns of deception; without it, the learning process is less aware of the overall narrative structure.

\subsection{Further Analysis}
To provide deeper insights into the omission-aware detection mechanism, we conduct further analyses from three perspectives: (\textit{i}) examining omission type distributions to understand prevalent deception patterns, (\textit{ii}) exploring LLM-based simulation strategies for scenarios without external news corpora, and (\textit{iii}) presenting representative cases demonstrating omission-based deception identification.

\subsubsection{Omission Type Analysis}

To understand the prevalent types of news omission, we analyze the omission information and intents identified by our model on real-world news data to summarize common omission types. Specifically, an LLM is prompted to categorize the identified omissions from our model's outputs into distinct types based on their working mechanisms.
We randomly sampled 500 news articles from each of the English and Chinese datasets. Through iterative summarization and consolidation, eight primary omission types are identified:
\textit{Contextual Omission} (omitting background information), \textit{Complexity Omission} (simplifying complex issues), \textit{Comparative Omission} (excluding comparative data), \textit{Impact Omission} (omitting potential consequences), \textit{Accountability Omission} (ignoring responsibility issues), \textit{Severity Omission} (minimizing perceived risks), \textit{Stakeholder Omission} (excluding diverse viewpoints), and \textit{Political Context Omission} (downplaying political motivations.)
The full prompt used, complete type definitions and representative examples can be found in the Appendix.

\begin{figure}[t]
\centering
    \begin{minipage}[b]{0.47\linewidth}
        \centering
        \includegraphics[width=.7\linewidth]{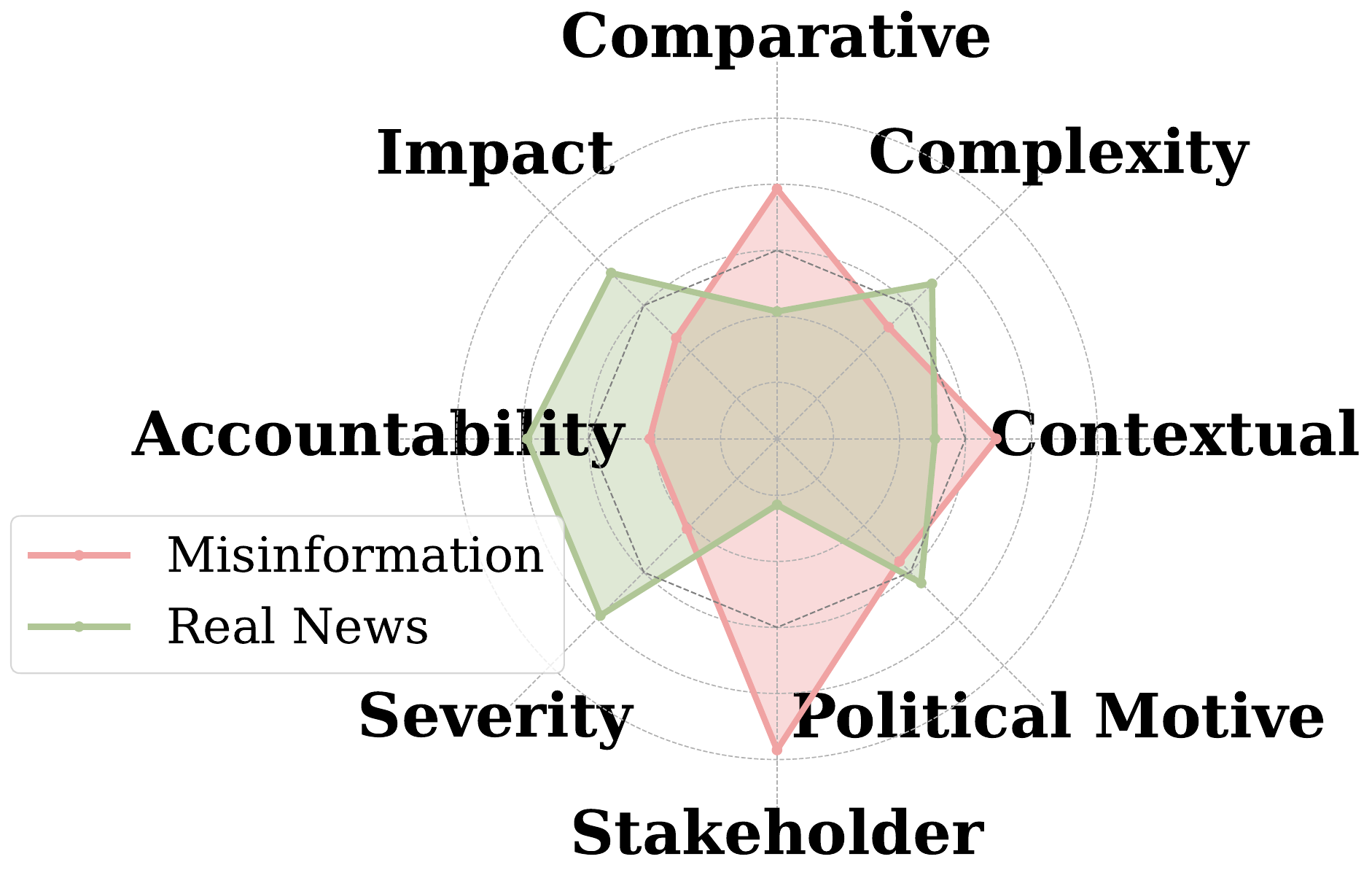}
    \end{minipage}
    \hfill
    \begin{minipage}[b]{0.47\linewidth}
        \centering
        \includegraphics[width=.7\linewidth]{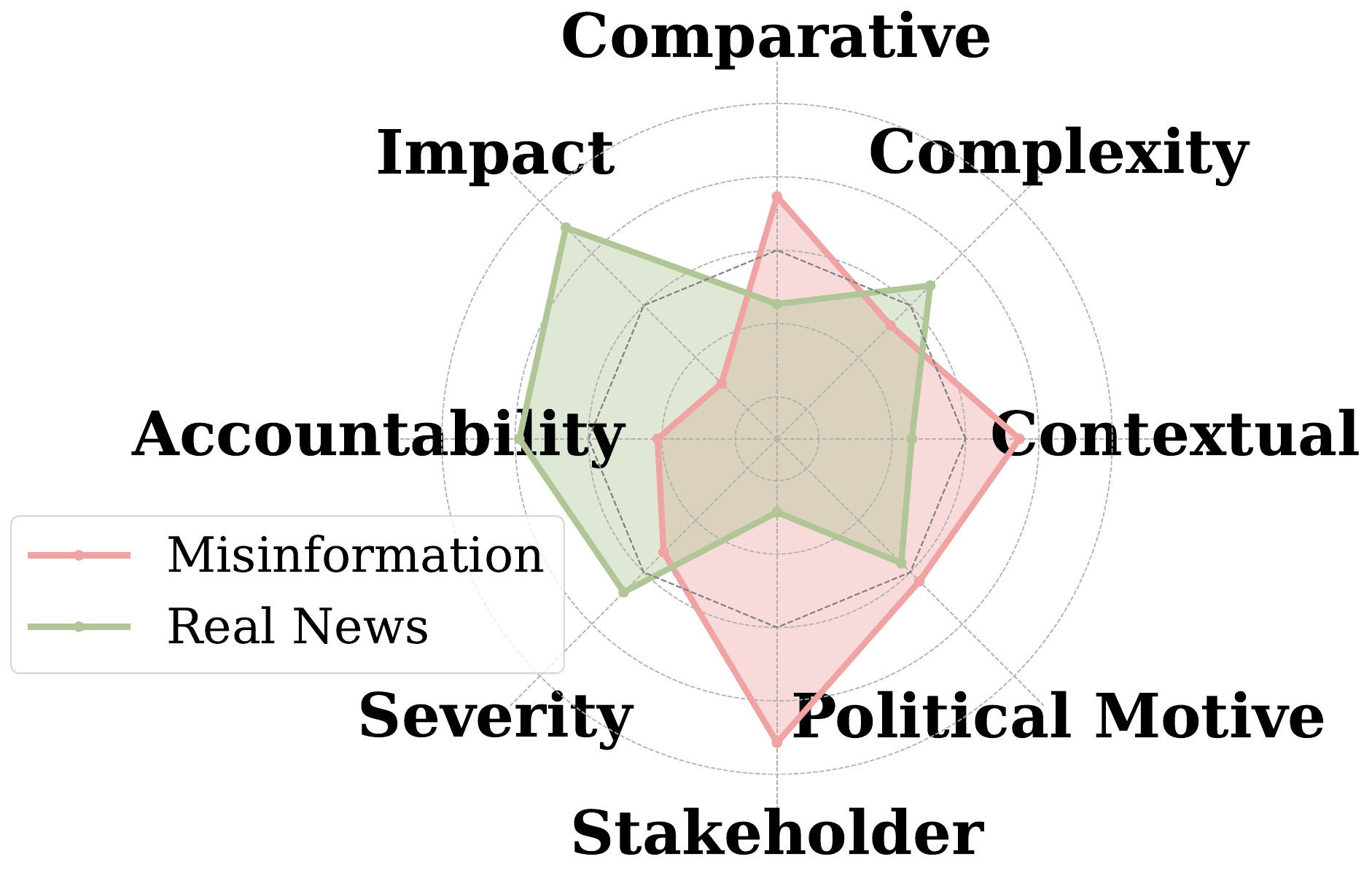}
    \end{minipage}
    \caption{Omission type distribution standardized using Z-scores in \textit{English} (left) and \textit{Chinese} (right) dataset.}
    \label{fig:omission_rader}
\end{figure}

We conducted LLM-assisted classification of the sampled news articles according to these omission types, with the distribution results shown in \figurename~\ref{fig:omission_rader}. The data was standardized using Z-scores to ensure uniform representation across all omission types.
The statistical analysis reveals distinct distributional patterns between real news and misinformation: misinformation exhibits higher rates of \textit{Comparative Omission} and \textit{Stakeholder Omission}, reflecting their tendency to manipulate statistical significance and suppress dissenting voices to support predetermined narratives. Conversely, real news shows a higher prevalence of \textit{Complexity Omission}. Interestingly, misinformation shows lower rates of \textit{Accountability Omission} and \textit{Severity Omission}, which conversely indicates that blame attribution and sensationalized severity descriptions are essential elements of many misinformation narratives.

\begin{table}[htbp]\small
  \centering
  \renewcommand{\arraystretch}{1.1}
  \caption{Performance of \modelname variants (with BERT as base detector) under simulation setting, compared with other LLM-based baselines. C$_{\text{token}}$ and C$_{\text{normed}}$ are token cost and normalized cost, respectively. \label{tab:sim}}
    \setlength\tabcolsep{4pt}
    \begin{tabular}{l ccc|rc}
    \toprule
    \multicolumn{1}{c}{\textbf{Method}}
    & \multicolumn{1}{c}{macF1} & \multicolumn{1}{c}{Acc} & \multicolumn{1}{c}{F1$_{\text{fake}}$}
    & \multicolumn{1}{|c}{C$_{\text{token}}$} & \multicolumn{1}{c}{C$_{\text{normed}}$}
    \\
    \midrule
    LLM
        & 0.5556  & 0.5779  & 0.6552
        & 103.2 & 0.07
        \\
    PCoT
        & 0.6508  & 0.6509  & 0.6481
        & 1488.5 & 1.00
        \\
    RAFTS
        & 0.6016  & 0.6049  & 0.6208
        & 1221.3 & 0.82
        \\
    \cmidrule{1-6} 
    \rowcolor{lightgrayv}\modelname 
        & \textbf{0.7530} & \textbf{0.7532} & \textbf{0.7519}
        & 951.9 & 0.64
        \\
    \; \textit{w/} Sim-Zero
        & 0.7341  & 0.7323  & 0.7396
        & 391.1 & 0.26
        \\
    \; \textit{w/} Sim-Rule
        & 0.7416  & 0.7427  & 0.7418
        & 570.6 & 0.38
        \\
    \bottomrule
    \end{tabular}%
  \setlength\dbltextfloatsep{-6pt}
\end{table}%

\subsubsection{LLM-based News Environment Simulation}

To demonstrate broader applicability when external news environments are unavailable, we explore using LLMs to simulate omitted information through two strategies: (\textit{i}) \textit{w/} Sim-Zero with direct prompting, and (\textit{ii}) \textit{w/} Sim-Rule using summarized omission types as guidance (detailed in Appendix).
For controlled comparison, we use BERT as the base misinformation detector across all variants.
We also evaluate efficiency by comparing token consumption with LLM-based baselines, reporting average token cost (C$_{\text{token}}$) and normalized cost ($C_{\text{normed}}$) scaled to the highest cost.

As shown in Table~\ref{tab:sim}, our approach (\textit{i}) demonstrates superior performance across diverse scenarios, validating the broader applicability of omission-aware detection beyond scenarios with external news corpora. (\textit{ii}) \modelname and its variants achieve superior performance with lower resource consumption, demonstrating that considering what is deliberately left unsaid is a new and cost-effective direction for misinformation detection.

\subsubsection{Case Analysis}
\begin{table}[t]
\caption{\label{tab:case}Case study illustrating the omitted information and omission intent extracted by \modelname.}
\centering
\renewcommand\arraystretch{1.2}
  \footnotesize
  \begin{tabular}{m{16cm}}
    \toprule
    \rowcolor{lightgrayv} \textbf{Target news}: Chicago city officials have adopted an official policy of protecting criminal aliens who prey on their residents.''
    \\
    
    \rowcolor{lightbluev} \textbf{Veracity label}: \textit{Misinformation} \quad \textbf{Source}: \textit{English}
    \\
    
    \midrule
    \textbf{Top $K$ contextual news}:\\
    \ding{172} Over threats to withhold public safety grant money, Chicago city officials have stated... \\
    \ding{173} Chicago To Sue Trump Administration Over Sanctuary City Funding Threat ``Chicago will not let our police... \\
    \ding{174} When The Police Are Criminals, Mexicans Have No One To Turn To ``The government doesn't listen.'' \\
    \ding{175} Truth is our National safety and economic is at risk ... \\
    
    \midrule
    \textbf{Omitted information captured by \modelname}:\\
    Chicago To Sue Trump Administration Over Sanctuary City Funding Threat ``Chicago will not let our police officers become political pawns in a debate,'' Mayor Rahm Emanuel said. \\

    \midrule
    \textbf{Omission intent captured by \modelname}:\\
    To obscure the context of the city's stance on immigration and law enforcement, which includes a defense of police autonomy. \\
    \bottomrule
  \end{tabular}
\end{table}

We present a representative case to demonstrate how \modelname works, showing how it facilitates misinformation detection through modeling omission-based deception.
As shown in \tablename~\ref{tab:case}, contextual news reveals important background information about Chicago's official stance on city policies and police autonomy, while this crucial information is deliberately omitted from the target news. Our method successfully detects this omission and infers the underlying intent, which reveals the deceptive strategy of obscuring the city's actual policy context to support the deceptive narrative. This omission-aware modeling effectively facilitates the detection of this misinformation case.

\section{Conclusion}
This paper introduces \modelname, the first omission-aware misinformation detection framework. By recognizing that deception operates not only through what is explicitly stated but also through what is deliberately omitted, \modelname addresses a critical yet underexplored dimension of news deception. 
Extensive experiments across bilingual datasets demonstrate the effectiveness and extensibility of our approach. Our analysis reveals common omission types and validates the value of omission-aware modeling for comprehensive deception understanding.

\normalem
\bibliography{sample-base}

\appendix

\section{Experimental Setup}

\subsection{Datasets}

We evaluate the proposed \modelname on two public datasets from \citep{sheng2022zoom}, which enables cross-lingual evaluation. 
The \textit{English} dataset comprises 6,483 verified posts from Twitter (now X) and fact-checking websites. 
The \textit{Chinese} dataset consists of 39,066 posts collected from Weibo, China's major social media platform. 
Both datasets are coupled with contemporary media coverage to establish comprehensive contextual news corpora, including 1,003,646 and 583,208 contemporaneous news articles for the English and Chinese datasets, respectively. 
For clarity, dataset statistics are shown in Table~\ref{tab:dataset}.

\begin{table}[htbp]\small
\centering
\renewcommand{\arraystretch}{1.01}
\setlength\dbltextfloatsep{-6pt}
    \setlength\abovecaptionskip{6pt}
    \setlength\belowcaptionskip{6pt}
\caption{\label{tab:dataset} Dataset statistics. These datasets encompass bilingual scenarios, facilitating comprehensive evaluation.}
  \setlength{\tabcolsep}{3pt}
  \begin{tabular}{lrrrrrrr}
  \toprule
  \multicolumn{1}{l}{\multirow{2}[4]{*}[0.3em]{\textbf{Dataset}}} & \multicolumn{2}{c}{\multirow{1}[1]{*}[0.2em]{\textbf{Train}}} & \multicolumn{2}{c}{\multirow{1}[2]{*}[0.3em]{\textbf{Validation}}} & \multicolumn{2}{c}{\textbf{Test}} & \multicolumn{1}{c}{\multirow{2}[4]{*}[0.1em]{\textbf{Total}}} \\
\cmidrule(lr){2-3}  \cmidrule(lr){4-5} \cmidrule(lr){6-7}
& \multicolumn{1}{l}{Fake} & \multicolumn{1}{l}{Real} & \multicolumn{1}{l}{Fake} & \multicolumn{1}{l}{Real} & \multicolumn{1}{l}{Fake} & \multicolumn{1}{l}{Real} \\
  \midrule
  \textbf{\textit{English}} 
        & 1,924 & 1,976 & 638 & 656 & 628 & 661 & 6,483\\
  \textbf{\textit{Chinese}} 
        & 8,992 & 8,787 & 4,923 & 5,131 & 5,608 & 5,625 & 39,066\\
  \bottomrule
  \end{tabular}%
  \setlength\textfloatsep{-16pt}
\end{table}%

\subsection{Implementation Details} 
Technically, our \modelname is designed to coordinate with any existing misinformation detection methods through flexible fusion strategies.
For approaches that produce news representations, our fusion scheme employs a Multi-Layer Perceptron (MLP) to generate fused prediction features from the combined representations. Specifically, we concatenate the feature representations from both our model and the baseline detector, then feed them through an MLP layer to produce the final fused prediction.
For LLM-based detection methods:
\begin{itemize}[leftmargin=*,itemsep=2pt,topsep=0pt,parsep=0pt]
    \item  \textit{LLM}, to validate the performance of LLM in the misinformation detection task, we prompt an LLM to make veracity judgments based on the provided news content.
    \item  \textit{PCoT} \citep{modzelewski2025pcot}, models misinformation from persuasion, integrating persuasion knowledge into the reasoning of LLMs for misinformation detection.
    \item \textit{RAFTS}~\citep{yue2024retrieval}, employs LLMs to construct contrastive arguments based on evidence, enabling nuanced reasoning for verification.
\end{itemize}
Our fusion approach performs post-prediction combination by integrating the prediction results from the base model with those from our model. The fusion is implemented using a tanh activation function, which helps normalize the combined predictions and improve stability.

To ensure fairness, all LLM-based methods employ the same model. Specifically, we prompt \emph{GPT-4o-mini}\footnote{https://openai.com/index/gpt-4o-mini-advancing-cost-efficient-intelligence/}, a well-recognized commercial LLM, to make a competitive comparison. 
The detailed prompt used for \textit{LLM} is listed in the following Prompt 1. 
\begin{tcolorbox}[title=Prompt 1:  Prompt for Veracity Judgment, boxrule=0pt, left=1mm, right=1mm, top=1mm, bottom=1mm, fontupper=\small]
    \textbf{System Prompt:} Given the following news piece, predict the veracity of this news piece. If the news piece is more likely to be fake, return 1; otherwise, return 0. Please refrain from providing ambiguous assessments, such as undetermined. \\
    \textbf{Context Prompt:} News: [\emph{$t$}]. The answer (Arabic numerals) is:
\end{tcolorbox}

For our proposed method \modelname, given the omission-aware graph $\mathcal{G}$, we elicit reasoning from an LLM $\mathcal{M}$ to analyze pairs of target and contextual news segments, inferring whether an omission relation exists and if so, \textit{why} information from contextual news segments is intentionally omitted.
The detailed prompt used is listed in Prompt 2. 
\begin{tcolorbox}[title=Prompt 2:  Prompt for Omission Intent Inference, boxrule=0pt, left=1mm, right=1mm, top=1mm, bottom=1mm, fontupper=\small]
    \textbf{System Prompt:} You are an AI annotator. You will be provided with two sets of news segments: [Target] and [Environment]. Your task is to detect potential omissions in the [Target] news content by comparing it against the contextual information provided in [Environment]. These omissions reflect intentional selective exclusion of information to better support the narrative.
    For each such pair, also analyze its omission intent (e.g., ``to prevent readers from thinking of the unreasonableness behind the overly high statistics''). Output in this format: {[Environment segment], [omission intent], [Target segment]}{\textbackslash n}
    \\
    Example:
    [The Start of Target] ``t1'' ``t2'' [The End of Target]
    {\textbackslash n}
    [The Start of Environment] ``e2'' ``e2'' [The End of Environment] {\textbackslash n}
    Your Answer:
    \{[``t1''], [omission intent], [ ``e1''] {\textbackslash n} [``t2''], [omission intent], [ ``e2'']\}
    \\
    \textbf{Context Prompt:} [The Start of Target]
    {\textbackslash n} [\emph{$t_1, t_2, ...$}] {\textbackslash n}
    [The End of Target]{\textbackslash n}
    [The Start of Environment]
    {\textbackslash n} [\emph{$e_1, e_2, ...$}] {\textbackslash n}
    [The End of Environment]{\textbackslash n}
    Your Answer:
\end{tcolorbox}

Using the above prompt, $\mathcal{M}(\cdot)$ returns omission intents between pairs of inter-source news segments through free-text descriptions ({\em e.g.}, ``to downplay the political motivations behind actions''), which are then encoded into edge attributes using a pre-trained language model.

We utilize AdamW as the optimizer, the batch size is 64, and the learning rate is $2e-5$. The feature embedding dimension is 256, and the MLP hidden sizes are $[128, 128]$.
We adopt \emph{bert-base-uncased} and \emph{bert-base-chinese}\footnote{https://huggingface.co/google-bert} as text encoders in Eq.6 for English and Chinese data, respectively.

\section{Hyperparameter Analysis} \label{sec:hyper_parameter}

\begin{figure}
    \centering 
    \includegraphics[width=.65\linewidth]{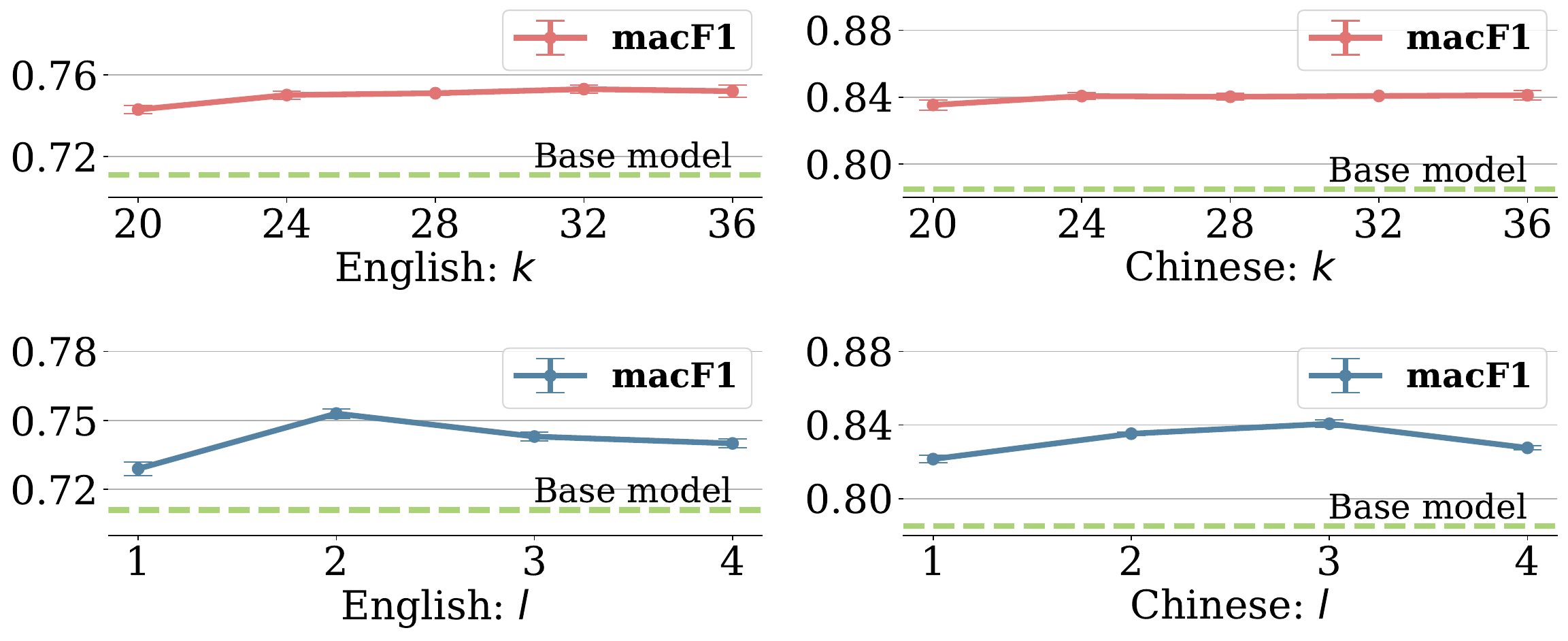}
    \caption{Impact of hyperparameters $k$ and $l$ on macro F1 scores (macF1) across English (left) and Chinese (right) datasets. The green dashed lines indicate the performance of BERT as the base model for controlled comparison.}\label{fig:hyperparameters}
\end{figure}

To understand the sensitivity of our model to key architectural choices, we conduct a hyperparameter analysis focusing on two critical parameters: the number of fine-grained contextual news segment nodes ($k$) and the number of stacked graph neural network layers ($l$). The results, illustrated in Figure~\ref{fig:hyperparameters}, provide insights into the optimal configuration and robustness of our approach.

For the impact of graph neural network layers ($l$), the analysis reveals a characteristic inverted-U relationship between the number of layers and model performance. Initially, increasing $l$ improves macro F1 scores across both English and Chinese datasets. This improvement can be attributed to the enhanced capacity of deeper networks to capture broader receptive fields and more complex relational patterns between news segments and omitted information. However, performance begins to degrade when $l$ continues to grow, likely due to the over-smoothing problem commonly observed in deep graph neural networks, where node representations become increasingly similar and lose discriminative power. The optimal configuration appears to be $l=2$ for English data and $l=3$ for Chinese, achieving the best balance between representational capacity and model stability.

The number of contextual news segment nodes $k$ is determined by selecting the top-K most similar items based on Eq.2. The sensitivity analysis for this parameter demonstrates remarkable robustness across different values of $k$. Performance remains relatively stable when $k$ varies, with only minor fluctuations observed in both English and Chinese datasets. This robustness suggests that our model's dynamic inference mechanism effectively adapts to varying amounts of contextual information, selectively focusing on the most relevant segments regardless of the total number available. The stable performance across different $k$ values indicates that the model can efficiently filter and prioritize informative segments while remaining resilient to potential noise introduced by less relevant contextual nodes. Specifically, considering the trade-off between computational cost and performance, we adopt $k=32$ for both datasets in our final configuration, which provides an optimal balance between model effectiveness and efficiency.

\section{Omission Type Analysis}

To understand the prevalent types of news omission, we conduct a systematic analysis of the omission information and intents identified by our model. This analysis aims to categorize and summarize common omission patterns that contribute to misinformation propagation. We randomly sampled 500 news articles from each of the English and Chinese datasets. Given the substantial volume of data, we adopted a batch-wise processing approach to ensure systematic and manageable analysis. 

Specifically, we divided the sampled articles into batches of $s=25$ news items each and employed an LLM to perform iterative categorization of the identified omissions.
For each batch, we prompted the LLM to analyze and abstract the omission type, categorizing them into distinct types based on their underlying mechanisms and characteristics. The detailed prompts used for this categorization are provided in Prompt 3. 

After processing all batches, we conducted a final consolidation phase that involved deduplication and hierarchical re-categorization to ensure comprehensive and non-overlapping omission types. The detailed prompts used for this process are provided in Prompt 4. 

Through this iterative summarization and consolidation process, we identified eight primary omission types that represent the most common patterns in news omissions:
\begin{itemize}
    \item \textbf{Contextual Omission}: Deliberately omitting relevant background information to obscure the broader implications of events or statements.
    \item \textbf{Complexity Omission}: Simplifying complex issues by omitting critical details that could provide a fuller understanding of the situation.
    
    \item \textbf{Comparative Omission}: Omitting relevant comparative data to exaggerate or downplay the significance of certain statistics or claims.
    
    \item \textbf{Impact Omission}: Omitting information about potential consequences or effects to create a misleading impression of significance.
    
    \item \textbf{Accountability Omission}: Ignoring issues of responsibility and accountability to simplify narratives surrounding controversial topics.
    
    \item \textbf{Severity Omission}: Omitting critical details that would highlight the seriousness of a situation, thereby minimizing perceived risks.
    
    \item \textbf{Stakeholder Omission}: Ignoring relevant viewpoints or public reactions to focus on a singular perspective, neglecting broader complexities surrounding a topic.
    
    \item \textbf{Political Context Omission}: Downplaying the political context or motivations behind actions to shape public perception.
\end{itemize}

\begin{tcolorbox}[title=Prompt 3: Batch-wise Omission Type Analysis, boxrule=0pt, left=1mm, right=1mm, top=1mm, bottom=1mm, fontupper=\small]
    \textbf{System Prompt:} You need to analyze the following omission intent analysis samples and summarize common omission types. {\textbackslash n} Example output format:{\textbackslash n}[Numerical Comparison Omission, Omitting relevant comparative data to exaggerate or downplay the importance of certain statistics]{\textbackslash n}[Background Information Omission, Deliberately omitting event background to prevent readers from understanding the complete situation]\\
    \textbf{Context Prompt:} Based on the following omission intent analysis samples, summarize the omission types that appear in this batch. Each sample contains: [segment with omission, omission intent, omitted information]{\textbackslash n}Please analyze these samples and output in the format: [Omission Type, Typical Intent Description].{\textbackslash n}
    Focus on identifying 3-8 distinct patterns in this batch:{\textbackslash n}[\emph{$b_1$}]
\end{tcolorbox}

\begin{tcolorbox}[title=Prompt 4: Final Omission Type Analysis, boxrule=0pt, left=1mm, right=1mm, top=1mm, bottom=1mm, fontupper=\small]
    \textbf{System Prompt:} Please remove duplicates, merge similar patterns, and output the final omission type summary.\\
    \textbf{Context Prompt:} Based on the following omission patterns summarized from different batches, merge similar types and extract the final 5-8 core omission patterns:{\textbackslash n}[\emph{$b_1, b_2, ...$}]{\textbackslash n}
\end{tcolorbox}

\begin{table}[htbp]\small
  \centering
  \renewcommand{\arraystretch}{1.1}
  \caption{Performance of \modelname variants on English dataset (with BERT as base detector) under simulation setting, compared with other LLM-based baselines. C$_{\text{token}}$ and C$_{\text{normed}}$ are token cost and normalized cost, respectively. \label{tab:sim2}}
    \setlength\tabcolsep{4pt}
    \begin{tabular}{l ccc|rc}
    \toprule
    \multicolumn{1}{c}{\textbf{Method}}
    & \multicolumn{1}{c}{macF1} & \multicolumn{1}{c}{Acc} & \multicolumn{1}{c}{F1$_{\text{fake}}$}
    & \multicolumn{1}{|c}{C$_{\text{token}}$} & \multicolumn{1}{c}{C$_{\text{normed}}$}
    \\
    \midrule
    LLM
        & 0.5556  & 0.5779  & 0.6552
        & 103.2 & 0.07
        \\
    PCoT
        & 0.6508  & 0.6509  & 0.6481
        & 1488.5 & 1.00
        \\
    RAFTS
        & 0.6016  & 0.6049  & 0.6208
        & 1221.3 & 0.82
        \\
    \cmidrule{1-6} 
    \rowcolor{lightgrayv}\modelname 
        & \textbf{0.7530} & \textbf{0.7532} & \textbf{0.7519}
        & 951.9 & 0.64
        \\
    \; \emph{w/} Sim-Zero
        & 0.7341  & 0.7323  & 0.7396
        & 391.1 & 0.26
        \\
    \; \emph{w/} Sim-Rule
        & 0.7416  & 0.7427  & 0.7418
        & 570.6 & 0.38
        \\
    BERT 
        & 0.7111    & 0.7135    & 0.7025
        & -  & -
    \\
    \bottomrule
    \end{tabular}%
  \setlength\dbltextfloatsep{-6pt}
\end{table}%

\section{LLM-based News Environment Simulation}

\begin{table*}[htbp]
\caption{Comparative case study of omission identification across different approaches. The table presents examples of omitted information identified by: ({left}) our full {\color{bettergreen}\modelname} method with contextual news environment, ({middle}) {\color{bluev}\emph{w/} Sim-Zero} (direct LLM prompting), and ({right}) {\color{yellowv}\emph{w/} Sim-Rule} (LLM guided by omission types), demonstrating the effectiveness of each strategy in generating relevant omissions and analyzing their intents.}\label{tab:omission_comparison_cases}

\centering
\renewcommand\arraystretch{1.}
  \footnotesize
  \begin{tabular}{m{5.45cm}m{5.45cm}m{5.45cm}}
    \toprule
    \rowcolor{lightgrayv}\multicolumn{3}{m{17.2cm}}{ \textbf{Target news}: This is not an earthquake, nor a disaster movie. This is the horrific scene when Beijing Fuwai Hospital opens in the morning, and patients rush to register. This kind of scene can only be seen in China among more than 200 countries in the world. This is the scene of Chinese people buying their lives with money, how sad and...}
    \\
    \rowcolor{lightbluev}\multicolumn{3}{m{17.2cm}}{\textbf{Veracity label}: \textit{Misinformation} \quad \textbf{Source}: \textit{Chinese}}
    \\
    \midrule
    \parbox[c]{5.45cm}{\textbf{Top $K$ contextual news}:\\
    \ding{172} An emergency doctor made six consecutive emergency calls on New Year's Eve, ... \\
    \ding{173} A 6-seater van was modified to hold 51 people. Is this a scene from a movie? No, this is a true story. \\
    \ding{174} The doctor said they had to make 6 trips a night and didn't even have time to eat ... 
    } 
    & \parbox[c]{5.45cm}{/}
    & \parbox[c]{5.45cm}{/}
    \\
    \midrule
    \parbox[t]{5.45cm}{\textbf{Omitted information ({\color{bettergreen}\modelname})}:\\
    \ding{172} An emergency doctor made six consecutive emergency calls on New Year's Eve, ... \\
    \ding{173} The doctor said they had to make 6 trips a night and didn't even have time to eat ...
    }
    & \parbox[t]{5.45cm}{\textbf{Omitted information ({\color{bluev} \emph{w/} Sim-Zero})}:\\
    \ding{172} Specific details about the hospital's capacity or the healthcare system's challenges. \\
    \ding{173} The sudden reason why patients rushed to register.
    }
    & \parbox[t]{5.45cm}{\textbf{Omitted information ({\color{yellowv}\emph{w/} Sim-Rule})}:\\
    \ding{172} Comparative data about healthcare situations in other countries. \\
    \ding{173} Details about the healthcare system's challenges in China. \\
    }
    \\
    \midrule
    \parbox[t]{5.45cm}{\textbf{Omission intent ({\color{bettergreen}\modelname})}:\\
    \ding{172} To conceal systemic strain and frontline efforts that contextualize the chaotic hospital scene. \\
    \ding{173} To suppress empathetic cues that might mitigate the outrage toward the system.
    }
    & \parbox[t]{5.45cm}{\textbf{Omission intent ({\color{bluev} \emph{w/} Sim-Zero})}:\\
    \ding{172} To arouse readers' emotional resonance and attention. \\
    \ding{173} To highlight the strain on medical resources and the anxiety of patients.
    }
    & \parbox[t]{5.45cm}{\textbf{Omission intent ({\color{yellowv}\emph{w/} Sim-Rule})}:\\
    \ding{172} To limit the reader's understanding of the uniqueness of the situation in China. \\
    \ding{173} To reduce a multifaceted issue to a single emotional reaction. \\
    }
    \\
    \hline
    \specialrule{0em}{0.5pt}{0.5pt}
    \hline
    \specialrule{0em}{0.pt}{3pt}
    
    \rowcolor{lightgrayv}\multicolumn{3}{m{17.2cm}}{ \textbf{Target news}: [Serving as a ``Guardian'' of Power Supply Safety] On December 16, the Gucheng Power Supply Station of the State Grid Nanjing Gaochun District Power Supply Company organized personnel to conduct a comprehensive inspection of substation lines and equipment west of Chanlinshan Village Road in Gucheng Subdistrict, acting as a ``Guardian'' of power supply safety ...}
    \\
    \rowcolor{lightbluev}\multicolumn{3}{m{17.2cm}}{\textbf{Veracity label}: \textit{Real} \quad \textbf{Source}: \textit{Chinese}}
    \\
    \midrule
    \parbox[c]{5.45cm}{\textbf{Top $K$ contextual news}:\\
    \ding{172} National Energy Administration: Multiple measures are being taken to address the tight power supply in parts of southern China \\
    \ding{173} Urge power grid companies to optimize their operation methods and strengthen ...
    } 
    & \parbox[c]{5.45cm}{/}
    & \parbox[c]{5.45cm}{/}
    \\
    \midrule
    \parbox[t]{5.45cm}{\textbf{Omitted information ({\color{bettergreen}\modelname})}:\\
    \ding{172} National Energy Administration: Multiple measures are being taken to address the tight power supply in parts of ... \\
    }
    & \parbox[t]{5.45cm}{\textbf{Omitted information ({\color{bluev} \emph{w/} Sim-Zero})}:\\
    \ding{172} Details and duration of supply constraints.
    }
    & \parbox[t]{5.45cm}{\textbf{Omitted information ({\color{yellowv}\emph{w/} Sim-Rule})}:\\
    \ding{172} The complexities of the power supply system and the specific challenges faced by the Gucheng Power Supply Station.
    }
    \\
    \midrule
    \parbox[t]{5.45cm}{\textbf{Omission intent ({\color{bettergreen}\modelname})}:\\
    \ding{172} To avoid people's concerns about insufficient power supply affecting service. \\
    }
    & \parbox[t]{5.45cm}{\textbf{Omission intent ({\color{bluev} \emph{w/} Sim-Zero})}:\\
    \ding{172} To affect public perception. \\
    }
    & \parbox[t]{5.45cm}{\textbf{Omission intent ({\color{yellowv}\emph{w/} Sim-Rule})}:\\
    \ding{172} To avoid overwhelming readers with technical details.
    }
    \\
    \bottomrule
  \end{tabular}
\end{table*}

To demonstrate the broader applicability of our approach when external news environments are unavailable, we explore the feasibility of using Large Language Models (LLMs) to simulate omitted information. This simulation-based approach provides an alternative solution for scenarios where comprehensive news corpora or external news information are not accessible.

We implement two distinct simulation strategies to generate omitted information: (\textit{i}) \emph{w/} Sim-Zero, which employs direct prompting without additional guidance, allowing the LLM to identify and generate omissions based solely on the input news content; and (\textit{ii}) \emph{w/} Sim-Rule, which leverages the eight omission types identified in our previous analysis as structured guidance to direct the LLM's generation process.

For controlled comparison across all simulation variants, we employ BERT as the base misinformation detector, ensuring that performance differences can be attributed to the simulation strategies rather than variations in the underlying detection model.

The detailed prompt used for omission simulation is presented below (the \emph{italic text} is only for \emph{w/} Sim-Rule):

\begin{tcolorbox}[title=Prompt 5: Prompt for LLM-based Simulation, boxrule=0pt, left=1mm, right=1mm, top=1mm, bottom=1mm, fontupper=\small] \label{prompt}
    \textbf{System Prompt:} You are an AI annotator. Given a set of news segments grouped by [Target], generate omitted information that reflects intentional omission to support specific narratives, and analyze omission intent for each pair.
    \emph{(e.g., 
    [Contextual Omission] omitting background information, [Complexity Omission] simplifying complex issues, [Comparative Omission] excluding comparative data, [Impact Omission] omitting potential consequences, [Accountability Omission] ignoring responsibility issues, [Severity Omission] minimizing perceived risks, [Stakeholder Omission] excluding diverse viewpoints, and [Political Context Omission] downplaying political motivations}
    {\textbackslash n}Output format: \{[Potential omitted information], [omission intent], [Target segment]\}
    \\
    \textbf{Context Prompt:} [The Start of Target Segments]{\textbackslash n} [\emph{$t_1,t_2,...$}]{\textbackslash n}[The End of Target Segments]{\textbackslash n}Your Answer:
\end{tcolorbox}

To assess the computational efficiency of our simulation approach, we conduct a comprehensive comparison of token consumption against LLM-based baseline methods. We report both the average token cost (C$_{\text{token}}$) and the normalized cost (C$_{\text{normed}}$), where the latter is scaled relative to the highest-cost method to provide intuitive cost comparisons across different approaches.

As shown in Table~\ref{tab:sim2}, our approach (\textit{i}) demonstrates superior performance across diverse scenarios, validating the broader applicability of omission-aware detection beyond scenarios with external news corpora. (\textit{ii}) \modelname and its variants achieve superior performance with lower resource consumption, demonstrating that considering what is deliberately left unsaid is a new and cost-effective direction for misinformation detection. The Sim-Rule variant incorporates the eight omission types (Contextual, Complexity, Comparative, Impact, Accountability, Severity, Stakeholder, and Political Context Omissions) as structured prompts to guide the LLM's generation process, ensuring more systematic and comprehensive omission identification compared to the zero-shot approach.

To illustrate how our approaches operate in practice, Table~\ref{tab:omission_comparison_cases} presents a comparative case study. As an English example has been provided in the main paper, we randomly select two cases from the Chinese dataset and translate them into English for accessibility.
As shown in the table, our full \modelname method ({\color{bettergreen}left column}) successfully captures potential omitted information from the contextual news environment and further analyzes the underlying omission intents. This capability proves instrumental in revealing the deliberate intentions behind selective information exclusion in news narratives, as well as benign omissions, thereby exposing potential manipulation strategies employed by misinformation sources.

For scenarios where external news environments are unavailable, our simulation approaches demonstrate promising effectiveness in generating plausible omitted information and corresponding intents. Notably, the Sim-Rule variant ({\color{yellowv}right column}), which incorporates the abstracted identified omission types as in-context guidance, exhibits superior pattern recognition capabilities compared to the direct prompting approach ({\color{bluev} middle column}). This in-context guidance enables the LLM to generate more systematic and comprehensive omissions aligned with abstracted deception patterns, highlighting the cross-instance commonality of omission strategies in misinformation. Understanding these omission-based deception patterns can effectively enhance the model’s detection capability.

The comparison reveals that while our full method with external news environment provides the most grounded and contextually accurate omission identification, the simulation strategies offer viable alternatives that maintain reasonable quality while addressing practical deployment constraints in resource-limited scenarios.

\section{Limitations and Future Work}
Despite the promising results demonstrated by our approach, several limitations warrant acknowledgment and suggest directions for future research.

\noindent\textbf{Dependency on External Knowledge Sources}. Our method relies on external news environments to identify omitted information. In scenarios where comprehensive news corpora or diverse news sources are unavailable, the effectiveness of our approach may be compromised. While we have explored LLM-based simulation as an alternative, this approach introduces potential biases inherent in the generative models.

\noindent\textbf{Omission Type Coverage}. While our analysis identifies eight primary omission types, the landscape of deceptive omissions may be more diverse and evolving. Future work should investigate whether additional omission patterns emerge in different domains ({\em e.g}., scientific news) or temporal contexts (e.g., during crisis).

\noindent\textbf{Future Research Directions}. Several promising avenues merit further investigation:
(\textit{i}) \textbf{\textit{Personalized Omission Awareness}}: Exploring how individual reader backgrounds, expertise levels, and information needs might influence the perception and impact of different omission types, leading to personalized mitigation and explanation systems.
(\textit{ii}) \textbf{\textit{Collaborative Verification}}: Developing frameworks that leverage crowdsourcing and expert knowledge to continuously validate and refine omission detection, creating more robust and community-driven misinformation detection systems.
(\textit{iii}) \textbf{\textit{Adaptive External Context and Generative Reasoning Fusion}}: In real-world settings, the external news environment can be noisy, incomplete, or misaligned. Future research could explore adaptive mechanisms to dynamically evaluate the utility of retrieved contextual information, filtering out irrelevant content while retaining useful signals. Furthermore, when external support is insufficient, learned omission patterns could be distilled to guide LLMs in hypothesizing plausible omitted information and intents, effectively simulating omission reasoning in low-context scenarios. This dual-pathway integration of external evidence and generative modeling might open up a more resilient and context-aware strategy for omission detection.

These directions build upon the foundation established by \modelname---the first omission-aware misinformation detection framework---highlighting how ``\textbf{Learning From Omission}'' offers a fundamentally novel and versatile paradigm. By demonstrating the feasibility and value of omission-aware modeling, \modelname opens new avenues for future research in trustworthy and interpretable misinformation mitigation solutions that can better serve the growing need in our increasingly complex media landscape.

\end{document}